# On Self-Regulated Swarms, Societal Memory, Speed and Dynamics

Vitorino Ramos[1], Carlos Fernandes[2] and Agostinho C. Rosa[2]

{[1]CVRM-IST, [2]LaSEEB-ISR}, Technical University of Lisbon (IST), {[1]TS, [2]TN 6.21},
Av. Rovisco Pais, 1, 1049-001, Lisbon, PORTUGAL
vitorino.ramos@alfa.ist.utl.pt, {cfernandes,acrosa}@laseeb.org

## Abstract

Wasps, bees, ants and termites all make effective use of their environment and resources by displaying collective "swarm" intelligence. Termite colonies - for instance - build nests with a complexity far beyond the comprehension of the individual termite, while ant colonies dynamically allocate labor to various vital tasks such as foraging or defense without any central decision-making ability. Recent research suggests that microbial life can be even richer: highly social, intricately networked, and teeming with interactions, as found in bacteria. What strikes from these observations is that both ant colonies and bacteria have similar natural mechanisms based on Stigmergy and Self-Organization in order to emerge coherent and sophisticated patterns of global foraging behavior. Keeping in mind the above characteristics we propose a *Self-Regulated Swarm* (SRS) algorithm which hybridizes the advantageous characteristics of *Swarm Intelligence* as the emergence of a societal environmental memory or cognitive map via collective pheromone laying in the landscape (properly balancing the exploration/exploitation nature of our dynamic search strategy), with a simple *Evolutionary mechanism* that trough a direct reproduction procedure linked to local environmental features is able to self-regulate the above exploratory swarm population, speeding it up globally. In order to test his adaptive response and robustness, we have recurred to different dynamic multimodal complex functions as well as to *Dynamic Optimization Control* problems, measuring reaction speeds and performance. Final comparisons were made with standard *Genetic Algorithms*, *Bacterial Foraging strategies*, as well as with recent *Co-Evolutionary* approaches. SRS's were able to demonstrate quick adaptive responses, while outperforming the results obtained by the other approaches. Additionally, some successful behaviors were found: SRS was able to maintain a number of different solutions, while adapting to unforeseen situations even when over the same cooperative foraging period, the community is requested to deal with two different and contradictory purposes; the possibility to spontaneously create and maintain different subpopulations on different peaks, emerging different exploratory corridors with intelligent path planning capabilities; the ability to request for new agents (division of labor) over dramatic changing periods, and economizing those foraging resources over periods of intermediate stabilization. Finally, results illustrate that the present SRS collective swarm of bio-inspired ant-like agents is able to track about 65% of moving peaks traveling up to ten times faster than the velocity of a single individual composing that precise swarm tracking system. This emerged behavior is probably one of the most interesting ones achieved by the present work.

## Environmental Dynamics

Most research in evolutionary (*EC*) and swarm intelligence (SI) computation focuses on optimization of static, non-changing problems. Many real-world optimization problems, however, are dynamic, and optimization is needed that are capable of continuously adapting the solution to a changing environment. In fact, many real-world problems are actually dynamic: new jobs have to be added to the schedule, machines may break down or wear down slowly, raw material is of changing quality, etc (*Branke*, [5,6]). If the optimization problem is dynamic, the goal is no longer to find the extrema, but to track their progression through the space as closely as possible. One method for achieving this is to evolve a function off-line that models the dynamics of the environment directly while another is to use the dynamics of the evolutionary -or Swarm Intelligent- process itself on-line to track the progression of the extrema. As described by *Angeline* [1] evolving a function that describes extrema dynamics is preferable when the dynamics can be modeled accurately off-line. When a single function cannot describe the dynamic accurately enough, an on-line approach using the implicit dynamics of the self-adaptive method is preferable. However, while evolutionary computation methods that evolve mutation variances perform well in static environments, it is not clear that these methods are beneficial when the gradient at each point is constantly in flux as in most dynamic environments [1], apart from significant, well-know and promising attempts. For these reasons, it seems appropriate to keep the suggestion of on-line methods, simultaneously attempting it with different computational paradigms such as those derived from Swarm Intelligence which allows for distributed real-time self-organization of solutions while maintaining strong adaptive capabilities. Following this new research path, we propose a *Self-Regulated Swarm* (*SRS*) algorithm which hybridizes the advantageous characteristics of Swarm Intelligence as the emergence of a societal environmental memory or cognitive map [31,10,27,26] via collective pheromone laying in the landscape (balancing the exploration/exploitation nature of the search strategy), with a simple evolutionary mechanism that through a direct reproduction procedure linked to local environment features is able to self-regulate the above exploratory swarm population, speeding it up globally [13]. Our present proposal is fully discussed in section II.

## Approaches

Dynamic Optimization (*DO*) problems, on a more abstract level can be characterized by several situations; this might mean that the optimization function, the problem instance, or some restrictions may change, and thus the optimum to that problem might change as well [6]. If any of these events are to be taken into account in the optimization process, we call the problem dynamic or changing. Since many approaches are possible, *Branke* and *Schmek* [5] suggested in 2002, a classification of *DO* problems, surveying it, while classifying a number of the most widespread techniques that have been published in the literature so far to make evolutionary algorithms (*EAs*), and adaptive approaches more generally, suitable for changing optimization problems. Most of these approaches could be grouped into one of the following three categories established by them. Cited references illustrate some examples: *React on Changes* - The EA is run in standard fashion, but as soon as a change in the environment has been detected, explicit actions are taken to increase diversity and thus facilitating the shift to the new optimum (*Cobb* [11], *Vavak* [36], *Angeline* [1], *Bäck* [2], *Grefenstette* [14]); *Maintaining Diversity throughout the run* - Convergence is avoided all the time and it is hoped that a spread-out population can adapt to changes more easily (*Grefenstette* [15], *DeJong* [24], *Huang* and *Rocha* [20,34]); and to conclude *Memory-based Approaches* - The EA is supplied with a memory to recall useful information from past generations, which seems especially useful when the optimum repeatedly returns to previous locations (*Dasgupta* [12], *Xu* [25], *Branke* [7]). Finally some recent proposals have been made using a Swarm Intelligent (SI) [4,23] approach to attempt to solve these dynamic problems. Generally, *Swarm Intelligence* can be regarded as the property of a system whereby the collective behaviors of (unsophisticated) entities interacting locally with their environment cause coherent functional global patterns to emerge. SI provides a basis with which it is possible to explore collective (or distributed) problem solving without centralized control or the provision of a global model [31,30]. These entities can be either regarded as bio-inspired ant-like agents in which self-organization occurs through trail formation via pheromone deposition and evaporation [4,31,13,16,33,32] giving rise to the well know *Ant Colony Systems* (*ACS*) and *Ant Colony Optimization* (*ACO*) algorithms by *Dorigo* et al. [4], or as physical particles embodied with direction, velocity and intrinsic memory for best global and local position [23,8,28,9,22] know as *Particle Swarm Optimization* (PSO) algorithms developed by *Kennedy*, *Eberhart* et al. [23]. Examples of these self-organizing approaches include the works by *Guntsch* and *Middendorf* [16], *Branke* [5], *Ramos*, *Fernandes* and *Rosa* [31,13], *Passino* et al. [29], *Middendorf* and *Schmeck* [17], *Carlisle* and *Dozier* [8,9], *Parrott* [28], *Janson* [22], among others [30].

## Self-Regulated Swarms

Many structures built by social insects are the outcome of a process of self-organization [31,30,32], in which the repeated actions of the insects in the colony interact over time with the changing physical environment to produce a characteristic end state [19]. A major mediating factor is stigmergy [35], the elicitation of specific environment-changing behaviors by the sensory effects of local environment changes produced by previous and past behavior of the whole community. Stigmergy is a class of mechanisms that mediate animal-animal interactions through artifacts or via indirect communication, providing a kind of environmental synergy, information gathered from work in progress, a distributed incremental learning and memory among the society. In fact, the work surface is not only where the constituent units meet each other and interact, as it is precisely where a dynamical cognitive map could be formed, allowing for the embodiment of emergent adaptive memory, cooperative learning and perception [31,10,27,26]. Constituent units not only learn from the environment as they can change it over time. Its introduction in 1959 by Pierre-Paul Grassé[1], made it possible to explain what had been until then considered paradoxical observations: In an insect society individuals work as if they were alone while their collective activities appear to be coordinated. The stimulation of the workers by the very performances they have achieved is a significant one inducing accurate and adaptable response (check applications in [33,32]). Keeping in mind these characteristics we will present a stigmergic self-regulated model to tackle the collective adaptation of a social swarm for dynamic tracking, based on real ant colony behaviors.

As mentioned above, the distribution of the pheromone represents the memory of the recent history of the swarm (his social cognitive map), and in a sense it contains information which the individual ants are unable to hold or transmit [31]. There is no direct communication between the organisms but a type of indirect communication through the *pheromonal* field. In fact, ants are not allowed to have any local memory and the individual's spatial knowledge is restricted to local information about the whole colony pheromone density. In order to design this behaviour, one simple model was adopted [10], and extended due to specific constraints of the present proposal, in order to deal with 3D dynamic environments.

---

[1] Grassé, P.P.: La reconstruction du nid et les coordinations inter-individuelles chez *Bellicositermes natalensis* et *Cubitermes sp*. La théorie de la stigmergie : Essai d'interpretation des termites constructeurs. *Insect Sociaux* (1959), 6, 41-83. The phrasing of his introduction to the term stigmergy is worth noting (translated to English in [19]): *The coordination of tasks and the regulation of constructions do not depend directly on the workers, but on the constructions themselves.* **The worker does not direct his work, but is guided by it**. *It is to this special form of stimulation that we give the name Stigmergy (**stigma** - wound from a pointed object, and **ergon** - work, product of labor = stimulating product of labor).*

As described by *Chialvo* and *Millonas* [10,27,26], the state of an individual ant can be expressed by its position $r$, and orientation $\theta$. Since the response at a given time is assumed to be independent of the previous history of the individual, it is sufficient to specify a transition probability from one place and orientation $(r,\theta)$ to the next $(r^*,\theta^*)$ an instant later. In a previous works by *Millonas* [27,26], transition rules were derived and generalized from noisy response functions, which in turn were found to reproduce a number of experimental results with real ants. The response function can effectively be translated into a two-parameter transition rule between the cells by use of a pheromone weighting function (Eq.1):

$$W(\sigma) = \left(1 + \frac{\sigma}{1+\gamma\sigma}\right)^\beta \quad (1)$$

This equation measures the relative probabilities of moving to a cite $r$ (in our context, to a cell in the grid *habitat*) with pheromone density $\sigma(r)$. The parameter $\beta$ is associated with the osmotropotaxic sensitivity, recognised by *E.O. Wilson* as one of two fundamental different types of ant's sense-data processing. *Osmotropotaxis*, is related to a kind of instantaneous pheromonal gradient following, while the other, *klinotaxis*, to a sequential method (though only the former will be considered in the present work as in [10]). Also it can be seen as a physiological inverse-noise parameter or gain. In practical terms, this parameter controls the degree of randomness with which each ant follows the gradient of pheromone. On the other hand, $1/\gamma$ is the sensory capacity, which describes the fact that each ant's ability to sense pheromone decreases somewhat at high concentrations. In addition to the former equation, there is a weighting factor $w(\Delta\theta)$, where $\Delta\theta$ is the change in direction at each time step, i.e. measures the magnitude of the difference in orientation. As an additional condition, each individual leaves a constant amount $\eta$ of pheromone at the cell in which it is located at every time step $t$. This pheromone decays at each time step at a rate $k$. Then, the normalised transition probabilities on the lattice to go from cell $k$ to cell $i$ are given by $P_{ik}$ (Eq. 2, [10]), where the notation $j/k$ indicates the sum over all the surrounding cells $j$ which are in the local neighbourhood of $k$. $\Delta_i$ measures the magnitude of the difference in orientation for the previous direction at time $t$-1. Since we use a neighbourhood composed of the cell and its eight neighbours, $\Delta_i$ can take the discrete values 0 through 4, and it is sufficient to assign a value $w_i$ for each of these changes of direction. *Chialvo* et al., used the weights of $w_0$ =1 (same direction), $w_1$ =1/2, $w_2$ =1/4, $w_3$ =1/12 and $w_4$ =1/20 (U-turn). In addition coherent results were found for $\eta$=0.07 (pheromone deposition rate), $k$=0.015 (pheromone

Table 1 – High level description of the *Self-Regulated Swarm* (*SRS*) algorithm proposed:

```
/* Initialization */
For all ants do
  Place ant at randomly selected site r
  e[ant]=1.0
End For
/* Main loop */
For t = 1 to t_max do
  For all ants do
    /* According to Eqs. 1 and 2 */
    Compute W(σ) and P_ik
    Move to a selected neighboring site not occupied by other ant
    /* According to Eq. 3 (section II) */
    Increase pheromone P_r at site r: P_r= P_r+[η+p(Δ[r]/Δmax)]
    /* Reproduction procedure */
    Compute n, the number of occupied surrounding cells
    If ant meets ant (i.e., n ≥ 1) then
      Determine P**(n)
      Compute reproduction probability P* = P**(n) [Δ(r)/Δmax]
      If real random [0, 1] < P* then
        Create one ant with e[ant]=1.0 and place it randomly on one of
              the free cells surrounding the main parent at site r
      End If
    End If
  End For
  Evaporate pheromone by K, at all grid sites
  For all ants do
    Decrease ant energy: e[ant]=e[ant]- Δe
    If e[ant] ≤ 0.0 then
      Kill that ant
    End If
  End For
  Print location of all agents
  Print pheromone distribution at all sites
End For
/* Values of parameters used in experiments */
k = 1.3, η = 0.07, β=3.5, γ=0.2, Δe=0.1,
p = 1.9, t_max = 100 or 400 time steps.
/* Constant values */
P**(0) = P**(8) =0, P**(4) = 1, P**(5) = P**(3) =0.75,
P**(6) = P**(2) =0.5, P**(7) = P**(1) = 0.25
/* Useful references */
Check [31], [13], [33], [10], [27] and [26].
```

evaporation rate), $\beta$=3.5 (osmotropotaxic sensitivity) and $\gamma$=0.2 (inverse of sensory capacity), where the emergence of well defined networks of trails were possible. Except when indicated, these values will remain in the following tests framework. As an additional condition, each individual leaves a constant amount $\eta$ of pheromone at the cell in which it is located at every time step $t$. Simultaneously, the pheromone evaporates at rate $k$, i.e., the pheromonal field will contain information about past movements of the organisms, but not arbitrarily in the past,

$$P_{ik} = \frac{W(\sigma_i)w(\Delta_i)}{\sum_{j/k} W(\sigma_j)w(\Delta_j)} \quad (2)$$

$$T = \eta + p\frac{\Delta[i]}{\Delta_{max}} \quad (3)$$

since the field *forgets* its distant history due to evaporation in a time $\tau \cong 1/k$. As in past works, toroidal boundary conditions are imposed on the lattice to remove, as far as possible any boundary effects (e.g. one ant going out of the grid at the south-west corner, will probably come in at the north-east corner). In order to achieve emergent and *autocatalytic* mass behaviours around specific extrema locations (e.g., peaks or valleys) on the *habitat*, instead of a constant pheromone deposition rate $\eta$ used in [10], a term not constant is included. This upgrade can significantly change the expected ant colony cognitive map (pheromonal field). The strategy follows an idea implemented earlier by *Ramos* [33], while extending the *Chialvo* model into digital image habitats, aiming to achieve a collective perception of those images by the end product of swarm interactions. The main differences to the *Chialvo* work is that ants, now move on a 3D discrete grid, representing the dynamic functions which we aim to study (see next section) instead of a 2D *habitat*, and the pheromone update takes in account not only the local pheromone distribution as well as some features of the cells around one ant. In here, this additional term should naturally be related with specific characteristics of cells around one ant, like their altitude ($z$ value or function value at coordinates $x,y$), having in mind our present aim. So, our pheromone deposition rate $T$, for a specific ant, at one specific cell $i$ (at time $t$), should change to a dynamic value ($p$ is a constant = 1.93) expressed by equation 3. In this equation, $\Delta_{max} = |z_{max} - z_{min}|$, being $z_{max}$ the maximum altitude found by the colony so far on the function *habitat*, and $z_{min}$ the lowest altitude. The other term $\Delta[i]$ is equivalent to (if our aim is to minimize any given landscape): $\Delta[i] = |z_i - z_{max}|$, being $z_i$ the current altitude of one ant at cell $i$. If on the contrary, our aim is to maximize any given dynamic landscape, then we should instead use $\Delta[i] = |z_i - z_{min}|$. Finally, notice that if our landscape is completely flat, results expected by this extended model will be equal to those found by *Chialvo* and *Millonas* in [10], since $\Delta[i]/\Delta_{max}$ equals to zero. In this case, this is equivalent to say that only the swarm pheromonal field is affecting each ant choices, and not the *environment* - i.e. the expected network of trails depends largely on the initial random position of the colony, and in trail clusters formed in the initial configurations of pheromone. On the other hand, if this environmental term is added a stable and emergent configuration will appear which is largely independent on the initial conditions of the colony and becomes more and more dependent on the nature of the current studied dynamic *landscape* itself. As specified earlier, the environment plays an active role, in conjunction with continuous positive and negative feedbacks provided by the colony and their pheromone, in order to achieve a stable emergent pattern, societal memory and distributed learning by the community [31,30,33].

**Reproduction procedure**

In addition to the above advantageous characteristics of *Swarm Intelligence* as the emergence of a societal environmental memory or cognitive map [10] via collective pheromone laying in the dynamic landscape, we hybridized it with a simple evolutionary mechanism that through a direct reproduction procedure linked to some local environmental features is able to self-regulate the above exploratory swarm population, speeding it up globally. The full *SRS* strategy adapted for dynamic extrema tracking then finally consists of using *Swarms with Varying Population Size* (SVPS) proposed and analyzed earlier by *Fernandes*, *Ramos* and *Rosa* [13]. This characteristic is achieved by allowing ants to reproduce and die through their evolution in the landscapes. To be effective, the process of variation must incorporate some kind of environmental pressure towards successful behavior, that is, ants that reach peaks/valleys must have some kind of reward, by staying alive for more generations – generating more offspring - or by simply having a higher probability of generating offspring at each time step. In addition, the population density in the area surrounding the parents must be taken into account during a reproduction event. When one ant is created (during initialization or by the reproduction process) a fixed energy value is assigned to it ($e[ant] = 1$). Every time step, ant's energy is decreased by a constant amount of $\Delta e$ (usually 0.1). The ant's probability of survival after a time step is proportional to its energy during that same iteration, which means that after ten generations (with $\Delta e = 0.1$), this and other ants will inevitably die ($e[ant] = 0$). Within these settings one ant that is, for instance, 7 iterations old, has a probability of 0.3 to survive through the current time step. Meanwhile, for the reproduction process, we assume the following heuristic: an ant (main parent) triggers a reproduction procedure if it finds at least another ant occupying one of its 8 surrounding cells (*Moore* neighborhood is adopted). The probability $P^*$ (Eq.4) of generating offspring – one child for each reproduction event – is computed in two steps (see table I).

$$P^* = P^{**}(n)\left[\frac{\Delta(r)}{\Delta_{max}}\right] \quad (4)$$

Table 2 – Reproduction Probability $P^{**}$ **x** $N$ Moore neighbors:

| $n$ Moore neighbors | Reproduction probability $P^{**}(n)$ |
|---|---|
| $n = 0$ or $n = 8$ | 0.00 |
| $n = 4$ | 1.00 |
| $n = 5$ or $n = 3$ | 0.75 |
| $n = 6$ or $n = 2$ | 0.50 |
| $n = 7$ or $n = 1$ | 0.25 |

First, the surrounding area is inspected in order to see if it is too crowded. Being $n$ the number of occupied cells around this ant, the probability to reproduce $P^{**}$ is set to the values shown in table 2. Notice that: 1) an ant completely surrounded by other ants, or isolated ($n=8$, $n=0$) do not reproduce; 2) the maximum probability is achieved when the area that ant is half occupied ($n=4$). After the probability $P^{**}$ is set to one of the previous values, the final probability is computed according to Eq.4, with the help of $\Delta(r)$ and $\Delta_{max}$ (similarly as in Eq. 3). This operation guarantees that any ant reaching the higher/lower peaks/valleys has more chance to produce offspring (notice that one ant in the higher/lower site has $P^{**}=1$ if $n=4$ and will reproduce for certain). If this ant passes the reproduction test (table I), then a new agent is created occupying one of his vacant cells around the main parent. None infant ants are allowed to be allocated in places where other ants are. Finally notice that when $n=0$ or $n=8$ no reproduction takes place and that higher/lower (maximization/ minimization) ants have more chance to reproduce.

### Shifting the extrema – past work results

One of the features of *Swarms with Varying Population Size SVPS* discussed in [13] was the ability to adapt to sudden changes in the roughness of the landscape. These changes were simulated by abruptly replacing one test function by another after the swarm reached the desired regions of the landscape. Another way of simulating changes in the environment consists on changing the task from minimization to maximization (or vice-versa). The swarm performance was convincing and reinforced the idea that the system is highly adaptable and flexible. Further tests using *SVPS* concluded that varying population size increases the capability of the swarm to react to changing landscapes [13]. Figure 1 shows *SVPS* trying to find the lower values of *Passino F1* [29] until $t=250$, and then searching for the higher values. Comparisons with fixed sized swarms and *Bacterial Foraging Optimization Algorithms* (*BFOA*, [29]) were made in [31].

## Dynamic Environment Testbed and Results

In order to compare the relative behavior of our present *Self-Regulated Swarms* (*SRS*) approach for on-line tracking of dynamic extrema, we followed diverse kind of test beds,

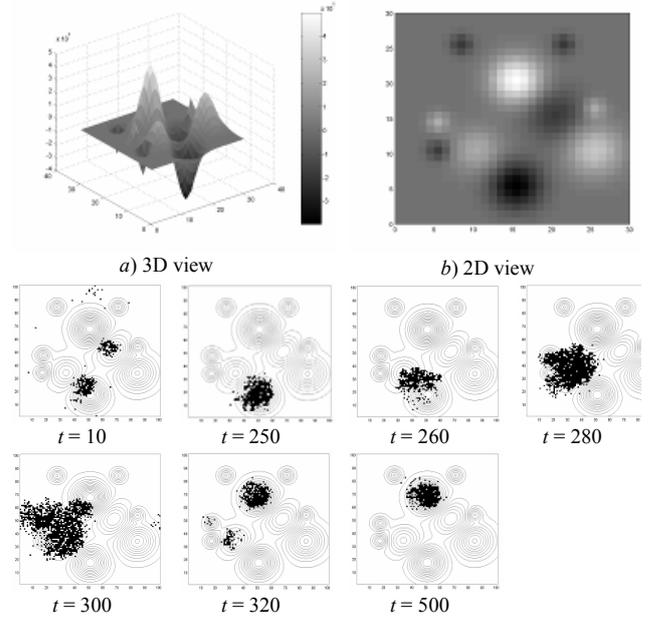

*a*) 3D view    *b*) 2D view

$t = 10$    $t = 250$    $t = 260$    $t = 280$

$t = 300$    $t = 320$    $t = 500$

Fig. 1. SVPS evolving on a complex multimodal function seen in *a-b*) [13,29,31]: the self-organized swarm emerges a characteristic flocking migration behavior between one deep valley (south region) and one peak (north region), surpassing in intermediate steps (*Mickey Mouse* shape at $t = 300$) some local optima. Over each foraging step, the population self-regulates. From $t=0$ to $t=250$ the swarm is induced to search the lowest valleys of the landscape. After $t=250$ the task changes (target peak moves to the north of the territory) and the swarm must find the higher values of the function. Check for detailed results in [13].

reflecting different dynamism characterization and benchmarks, which have been detailed described in [1,3,5,6], namely according to: *dynamic type*, *severity and speed*, as well as the application of the current self-organized algorithm to *dynamic optimal control* problems.

### Dynamic type tests

Over this specific test bed, dynamic environments were obtained by translating the *Ackley* complex multimodal base function (fig.2) along a number of distinct linear and circular temporal trajectories. This setup allows complete control of the dynamics and generates simple yet non-trivial dynamic functions with know properties. The *Ackley* function [30] (Eq.5) is considered as a minimization problem. Originally this problem was defined for two dimensions, but the problem has been generalized to $N$ dimensions [3]. Formally, this problem can be described as finding a string $x_i=\{x_1, x_2, …,x_N\}$, under the domain $x_i \rightarrow$ (-32.768, 32.768), that minimizes equation 5. In order to define an instance of this function we need to provide the dimension of the problem (in here, $n=2$). The optimum solution of the problem is the vector $u = (0,...,0)$ with $F(u)=0$.

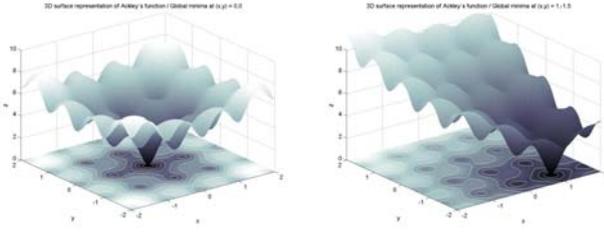

Fig. 2. The *Ackley* complex multimodal function seen from different perspectives and with respective global minimum $z_{min}=0$ at $B(0;0)$ (left) and $C(1;1.5)$ (right) over the domain $x,y \to [-2,2],[-2,2]$ with 16 valleys.

$$F\left(\vec{x}\right) = -20.\exp\left(-0.2\sqrt{\frac{1}{n}.\sum_{i=1}^{n}(x_i - a_i)^2}\right) - \exp\left(\frac{1}{n}\sum_{i=1}^{n}\cos(2\pi.(x_i - a_i))\right) + 20 + e \quad (5)$$

For our specific tests we have made use of the domain $x,y \to [-2.0,2.0],[-2.0,2.0]$, containing 16 valleys. In order to change it over time, we have defined a target path line over our domain, from northwest (-2,2) to southeast (2,-2). Figure 2, shows one typical minimal target point (*B*) along the line, *A*, *B* and *C*: *A*(-1.5;1.0), *B*(0.0;0.0) and *C*(1.0;-1.5). To introduce the dynamics into the problem, the base function minimal target point starts to move from *B*, moving continuously to southeast (passing *C*). Arriving at the southeast corner, the target point then moves to the northwest corner (since our *habitat* is toroidal), then passing *A*, and *B* again, over and over again has times passes, for several specific test speeds. Note that in equation 5, $a_i$ defines the point coordinates where the *Ackley* function has his minimal target point, thus introducing a more severe change than translation itself, since around that minimal point all the function changes differently depending on where this minimal point precisely is. The *SRS* algorithm was then tested for different test speeds: $v=0$ (static environment), $v=0.5$, $v=1$, $v=1.5$, $v=2$, $v=3$, $v=5$ and $v=10$. In order to have an idea of the severity of this parameter, for instance at $v=2$, the target valley is traveling to southeast 2 habitat cells at each *t* time step starting in the *B*(0.0;0.0) mid-point, while each *SRS* ant can only move one cell for each time step *t*. *SRS* parameter values were the usual ones as indicated in table 1. Somehow an emergent short path strategy is followed by the swarm, splitting in different clouds when needed for test speeds equal to $v=2$ or above that value (figs. 3b,3c). For dramatic values of speed however ($v=5$), the swarm is constantly and properly following the target but due to the value of severe speed, he is loosing the run. In order to overcame this lack of control the swarm itself self-regulates its population, and for dramatic speed tests of $v=10$ (fig.3a) or above, he explodes its population, trying to collect much spatial information as possible.

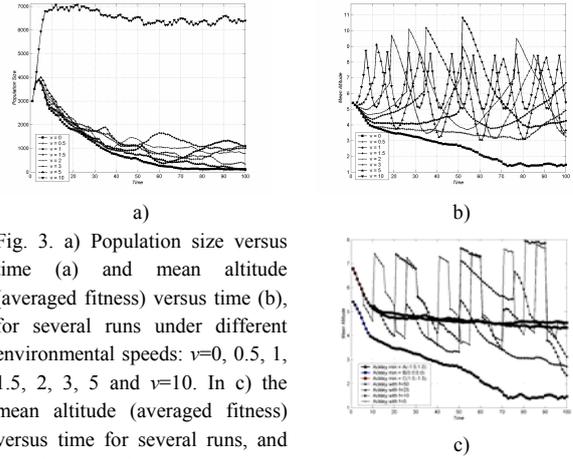

Fig. 3. a) Population size versus time (a) and mean altitude (averaged fitness) versus time (b), for several runs under different environmental speeds: $v=0$, 0.5, 1, 1.5, 2, 3, 5 and $v=10$. In c) the mean altitude (averaged fitness) versus time for several runs, and for different landscape change updated frequencies: $uf=50$, 25, 10 and 5.

Figure 3 can help us understand more profoundly the swarm behavior facing these different speeds. Fig. 3a shows the population size as time passes. Every test started with a number of ant-like agents equal to ⅓ of the habitat size (100 x 100 cells) as in [10,33]. We see that generally for speed tests up to $v=5$, the population more or less self-regulates under a common value, indicating a possible good tracking behavior accompanied by agent's specialization (exploitation nature). On the other hand, figure 3b shows what happens to mean agent's altitude over the *Ackley* landscape (*F* value on Eq. 5) as time go by. For speed tests up to $v=2$, the swarm can without difficulty track the extrema. For speed tests above that value, the behavior is more oscillatory, since at the toroidal borders the function assumes - due to simple and normal constraints in our representation - abrupt values from one border cell to their neighbors on "the other side". While the extrema is reasonably well tracked, ant-like agents very near the minimal extrema (near borders) could easily have high altitude "on the other side". Thus, even if the extrema is well tracked, mean altitude values achieve a visible increment. In addition, the swarm reaction speeds are undeniably very fast and the mean altitude quickly decreases often to values equal of those obtained with test speeds less or equal than $v=2$. This is particularly evident if we conduct a different and more dramatic test. In figure 3c we have updated the changes in the environment for different updated frequencies $uf$. For $uf=5$ (worst case) the base function minimal target point changes abruptly from *B* to *C*, from *C* to *A*, and from *A* to *B*, etc, at every 5 time steps. We see that even for this dramatic $uf$ values, the swarm as a whole can still obtain performances, which in some cases outperform those obtained for static environments (bold curves). Reactions speeds are also

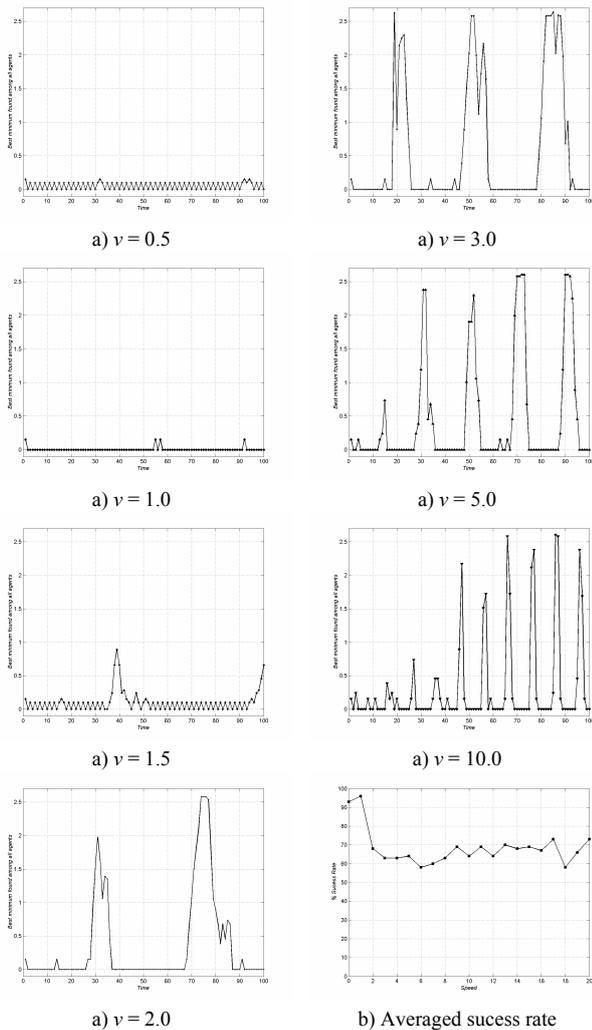

a) $v = 0.5$   a) $v = 3.0$
a) $v = 1.0$   a) $v = 5.0$
a) $v = 1.5$   a) $v = 10.0$
a) $v = 2.0$   b) Averaged sucess rate

Fig. 4. a) Best minimum value found by *SRS* among all agents at each time step (zero is equivalent to full performance or the perfect capture of the right extrema). Plots are for different test speeds. In b) the averaged success rate (10 runs for each test speed). The success rate indicates the number of times the colony as a whole was able to track the right target over a run of 100 time steps, at each speed.

quite high, generally in the range order of four to nine time steps. The *SRS* performance can also be measured and analyzed taking in account the best minimum value found among all agents at each time step (zero is equivalent to full performance or the perfect capture of the right extrema). In figure 4 we have plotted these values for different test speeds. As expected, as speed increases the swarm passes more time away from the right extrema or close to it. The swarm however is able to capture the right extrema in perfection, for the most part of the run-time (the oscillatory behavior is again mainly due to our toroidal borders). From these results we can obtain the *SRS* success rate (fig. 4b), defined as the number of times the colony as a whole was able not only to track, but to capture the right target over a run of 100 time steps, for each test speed. Remarkably the swarm is generally able to capture the right perfect extrema around 65% of the time, up to environmental speeds of 10 times the speed of a single ant-like agent or even for speeds greater than those values. Finally, very similar results were found for circular dynamics. Random dynamics were not tested.

**Speed tests and severity**

*Angeline* [1] as well as *Bäck* [2,3], *Branke, Schmeck* [5,6,7] and others, studied not only how adaptive algorithms behave in drastic changing environments (as in the preceding subsection), but also on how the degree of these changes affect different approaches. The question is important, since can help us understand on how strongly the system changes are and how severe this change is going to constrain the search nature of an algorithm. For these types of tests as well as for comparison purposes we have decided to implement a test function used recently by *Huang* and *Rocha* [20,34,21] with a Co-Evolutionary agent-based model of Genotype Editing (*ABMGE*) [34]. This testbed is a dynamic version of the modified *Schaffer's F7* function studied in [21]. Several sketches of this multimodal function for different parametric values (different instances in time) are illustrated in fig. 5. Being *s* representing a parameter to set the severity and $X_i = x_i + \delta(t)$, with $-1 \leq x_i \leq 1$ for $i = 1,2$, a possible test problem [20] can be described by Equation 6. In here, *X* and *Y* axis represent the index of the sample points in parameters $x_1$ and $x_2$ that are used to compute $f(x)$, which is then plotted on the *z* axis, being our aim to maximize it. The example uses linear dynamics with severity *s* (Eq.7).

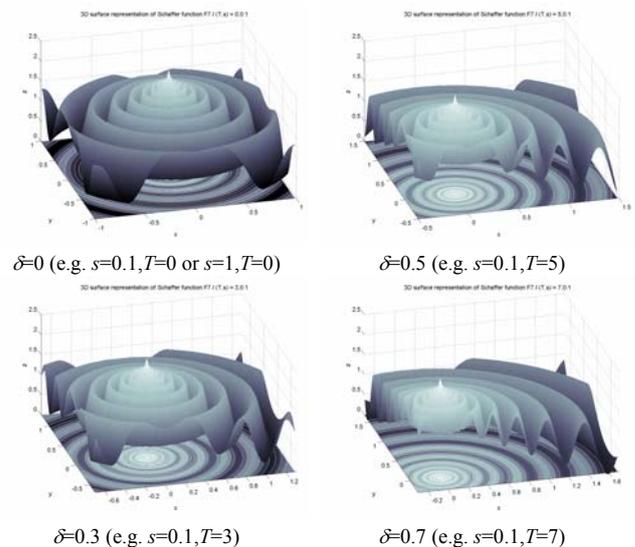

$\delta=0$ (e.g. $s=0.1, T=0$ or $s=1, T=0$)   $\delta=0.5$ (e.g. $s=0.1, T=5$)
$\delta=0.3$ (e.g. $s=0.1, T=3$)   $\delta=0.7$ (e.g. $s=0.1, T=7$)

Fig. 5. Sketches from the *Schaffer F7* complex multimodal function seen for different values of *T* related to the severity tests (here, $s=0.1$).

$$f\left(\vec{X}\right) = 2.5 - \left(X_1^2 + X_2^2\right)^{0.25} \left[\sin^2\left(50 \cdot \left(X_1^2 + X_2^2\right)^{0.1}\right) + 1\right] \quad (6)$$

$$\begin{aligned}\delta(0) &= 0, \\ \delta(T) &= \delta(T-1) + s\end{aligned} \quad (7)$$

As explained in [20], note that $T$ is used as index for the environmental state; whenever the environment changes (e.g., every 50 generations in [20]), $T$ is increased by 1. With these settings, $f(x)$ has an optimal value of 2.5 among all ranges of $s$, here tested (this happens for $s=0.1$). This equation (7) will also be used for the dynamic control test function studied in the next sub-section. In here we tested the *SRS* algorithm against severity values of $s=0.1$, $s=0.2$, $s=0.3$, $s=0.5$, $s=1.0$, and $s=1.5$ (for a constant updated frequency, $uf = 50$), as well as for different updated frequency's $uf = 50, 25, 10$ and 5 (worst dramatic case), for a constant severity of $s=1$. *SRS* parameter values were the usual ones as indicated in table 1. Figure 6 displays the results achieved by *Huang* [20]; averaged best-so-far performance for the *Schaffer's F7* (Eq. 6) dynamic problem, with traditional genetic algorithms as well as with the co-evolutionary RNA editing model (*ABMGE*) [34], for a single test with $s=0.1$ and $uf=50$. As outlined by the authors, these results are encouraging since the co-evolutionary *ABMGE* model consistently outperforms the traditional GA in tracking the extrema. Meanwhile, figure 6b displays the results achieved by *SRS* against not only $s=0.1$ as well as for other values of severity $s$ (all with $uf=50$). We see that the self-regulated swarm is not only able to track perfectly the correct extrema, nearly at all iterations for $s=0.1$, as achieves similar results for increasing values of severity. Please note that when $s$ increases, equation 7 returns a different domain for our test dynamic function (different sample points for different $T$), thus the right respective extrema have optimal values less and obviously different than those present for $s=0.1$ ($z=2.5$). Meanwhile, figure 6c displays the *SRS* population size over the same tests. Since generally, the optimal peak form is the same (due to translation), for different severities (which did not happen in earlier tests), we now see *SRS* self-regulating its population size similarly for different test runs (*SRS* tend to converge his population to values equal of those optimal and near optimal areas; circular concentric regions in fig. 5). As usual, *SRS* tends also to use more exploratory resources as severity increases. Finally, we tested *SRS* against different updated frequencies. Respective results of performance and population size can be seen in figures 6d and 6e. We see that even for the worst case scenario ($uf=5$ and $s=1$), *SRS* swarms are able to track the correct extrema in the majority of the test run-time. Naturally, there are oscillatory behaviors when the frequency updates, but again the swarm reaction speed is quite fast: in the order of ten to fifteen time-steps. As usual, at these kind of critical periods (environmental phase transitions) the swarm explodes his population (as in fig. 3a), in order to overcome the abrupt changes in the landscape, using intelligently more resources to explore the novel sudden *habitat*. When again the extrema is correctly tracked after ten to fifteen time-steps, *SRS* self-regulates the number of their agents, downsizing it. Indeed, after the initial shock the resource economy is now so dominant and wise, that for certain cases the population decreases about nine times their earlier maximum size, settling down only at the correct peak.

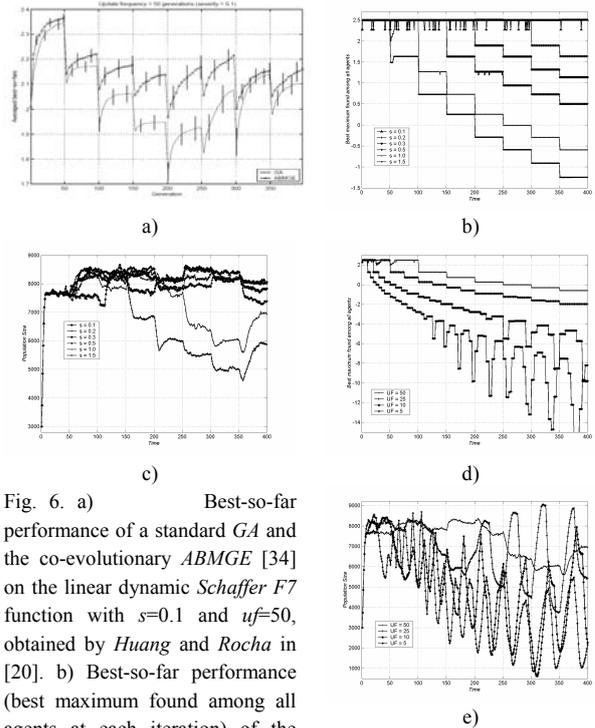

Fig. 6. a) Best-so-far performance of a standard *GA* and the co-evolutionary *ABMGE* [34] on the linear dynamic *Schaffer F7* function with $s=0.1$ and $uf=50$, obtained by *Huang* and *Rocha* in [20]. b) Best-so-far performance (best maximum found among all agents at each iteration) of the self-regulated swarm *SRS* approach against different severity values $s=0.1$, $s=0.2$, $s=0.3$, $s=0.5$, $s=1.0$, and $s=1.5$ ($uf=50$). c) *SRS* swarm population size at each iteration, for tests in figure b). d) Best-so-far performance (best maximum found among all agents at each iteration) of the self-regulated swarm *SRS* approach against different updated frequencies $uf=50, 25, 10$ and 5 ($s=1$). e) *SRS* swarm population size at each iteration, for d) tests.

**Dynamic Optimal Control problems**

Optimal Control Theory aims to determine the control signals that will cause a process to satisfy the physical constraints and at the same time minimize (or maximize) some performance criterion. Practically, this kind of nonlinear and multiple local optimal problems (generally) appear in all engineering and science fields, and have been well studied from both theoretical and computational perspectives [20]. Once more, for comparison purposes we adopt an earlier artificial optimal control problem designed in [21], further studied and reported in [20]. The constraints of the artificial optimal control problem are (Eq. 8):

$$\frac{d^2z(t)}{dt^2} + \sin(z(t))\cdot\frac{dz(t)}{dt} + \sin(t)\cdot\cos(z(t))\cdot z(t)^3 = \sin(t)u_1^2 + \cos(t)u_2^2 + \sin(t)u_1.u_2,$$
$$z(t_o) = 2, z'(t_0) = 2, t \in [0,1] \quad (8)$$

where $U_i = u_i + \delta(t)$, with $-5 \le u_i \le 5$ for $i = 1,2$. As in [20], two examples are studied in this subsection that use linear dynamics with severity $s=0.1$ and $s=1$, according to equation 7 in the previous subsection, respectively. The goal is to maximize $z(t_f)^2$ by searching for two control variables, $u_1$ and $u_2$ ($-5 \le u_1, u_2 \le 5$). Sketches of this dynamic functions for $\delta$=0, 2, 3, 4, 6 and 7 are respectively illustrated in figure 7. $X$ and $Y$ represent the index of sample points in parameters $u_1$ and $u_2$, that are used to

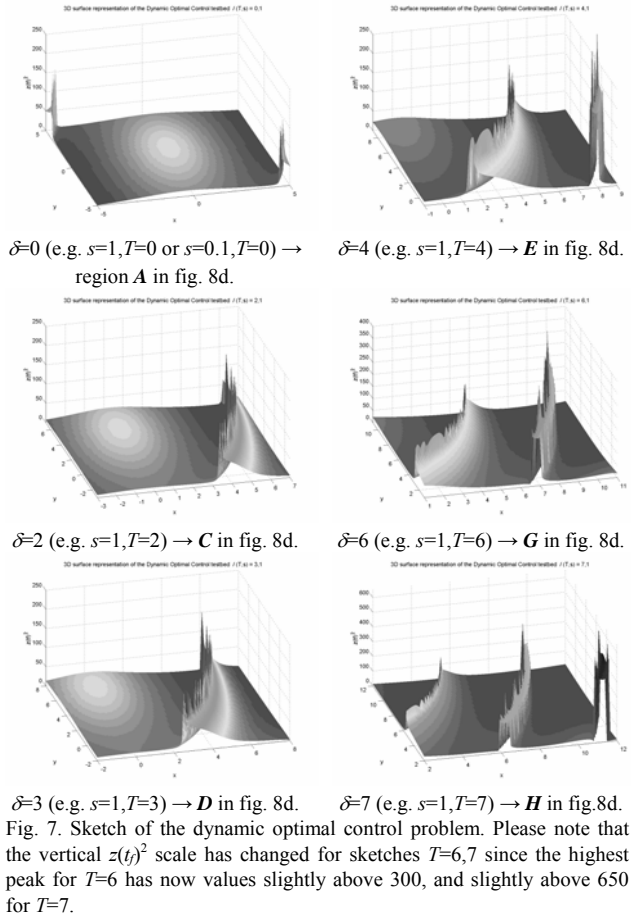

$\delta$=0 (e.g. $s$=1,$T$=0 or $s$=0.1,$T$=0) → region $A$ in fig. 8d.
$\delta$=4 (e.g. $s$=1,$T$=4) → $E$ in fig. 8d.
$\delta$=2 (e.g. $s$=1,$T$=2) → $C$ in fig. 8d.
$\delta$=6 (e.g. $s$=1,$T$=6) → $G$ in fig. 8d.
$\delta$=3 (e.g. $s$=1,$T$=3) → $D$ in fig. 8d.
$\delta$=7 (e.g. $s$=1,$T$=7) → $H$ in fig.8d.

Fig. 7. Sketch of the dynamic optimal control problem. Please note that the vertical $z(t_f)^2$ scale has changed for sketches $T$=6,7 since the highest peak for $T$=6 has now values slightly above 300, and slightly above 650 for $T$=7.

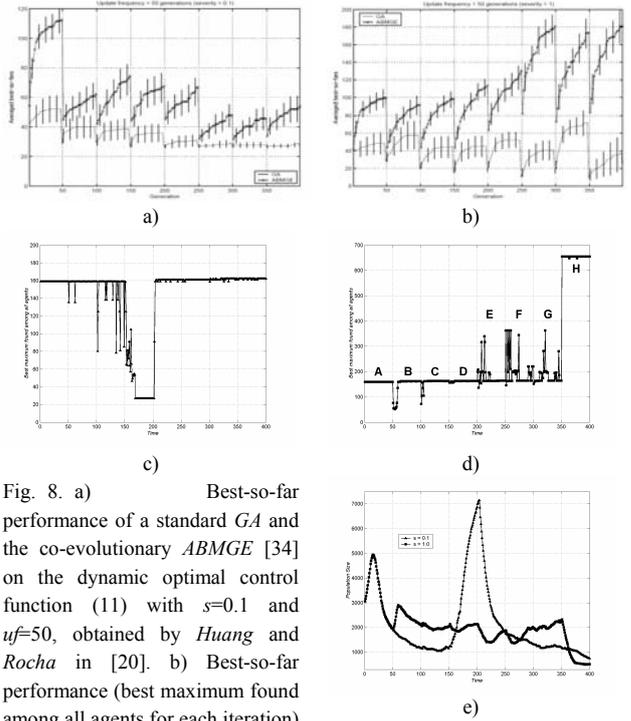

a)
b)
c)
d)
e)

Fig. 8. a) Best-so-far performance of a standard *GA* and the co-evolutionary *ABMGE* [34] on the dynamic optimal control function (11) with $s$=0.1 and $uf$=50, obtained by *Huang* and *Rocha* in [20]. b) Best-so-far performance (best maximum found among all agents for each iteration) of the self-regulated swarm *SRS* approach on the dynamic optimal control function (11) with $s$=0.1 and $uf$=50. c) Best-so-far performance of a standard *GA* and the co-evolutionary *ABMGE* [34] on the dynamic optimal control function (11) with $s$=1 and $uf$=50, achieved by *Huang* and *Rocha* in [20]. d) Best-so-far performance (best maximum found among all agents for each iteration) of the self-regulated swarm *SRS* approach on the dynamic optimal control function (11) with $s$=1 and $uf$=50. Regions A,B, …, H correspond to regions in figure 18. e) Population size as time passes, of the self-regulated swarm *SRS* approach on the dynamic optimal control function (11) with $s$=0.1 and $s$=1 ($uf$=50).

compute $z(t_f)^2$, which is then plotted on the Z-axis. In order to compute $z(t_f)^2$, the $2^{nd}$ order differential equation (*DE*) in (8) must be converted into an equivalent system of two $1^{st}$ order ordinary differential equations. The standard optimal control problem should now be represented by *max* $J(u_1, u_2) = z(t_f)^2$, with $t_f = 1$, $z(t_0)$=2, $y(t_0)$=2, $dz(t)/dt = y(t)$, and $dy(t)/dt$ given by Eq. 8. In order to numerically solve it for the different range values of $\delta$ (see Eq. 7), we have made use of a MATLAB version, working in reverse communication, of the Ordinary Differential Equation (*ODE*) solver DOPRI853 developed and coded in FORTRAN by *Hairer* et al. [18]. As in [20] we have completed tests for 400 iterations, with $s$=0.1 and $s$=1. Every 50 generations ($uf$=50), $T$ (Eq. 7) is increased by 1. We have used *SRS* values of $\Delta e$=0.01 (remaining parameter values were the usual ones as indicated in table 1). Figures 8a and 8b, respectively show traditional *GA* and *ABMGE* results for $s$=0.1 and 1, achieved by *Huang* and

*Rocha* [20,34]. They show that RNA editors [34], provide adaptive advantage for the ABMGE approach in tracking the extrema, while comparing it with traditional *GA*s. On the other hand, figures 8c and 8d, respectively show *SRS* results for $s=0.1$ and 1, achieved by our self-regulated swarm, clearly outperforming not only traditional *GA*s as the *ABMGE* model when facing dynamic environments. An exception however, should be made for $t \rightarrow [160,200]$ over the $s=0.1$ severity test, were *SRS* population exploded, due to sudden upcoming novel peak values over the domain. Fig. 8e shows *SRS* population size as time passes, for the $s=0.1$ and $s=1$ tests. Generally, the standard *GA* and *ABMGE* does not have time, within 50 generations to exploit better peaks. As we can see from these pictures the SRS reaction speed is generally higher, maintaining good (if not perfect) results for a large range of different time-steps and changing conditions.

## Conclusions

In order to overcome difficult dynamic environment extrema tracking, we have proposed a *Self-Regulated Swarm* (*SRS*) algorithm which hybridizes the advantageous characteristics of Swarm Intelligence as the emergence of a societal environmental memory or cognitive map via collective pheromone laying in the landscape (balancing the exploration/exploitation nature of the search strategy), with a simple evolutionary mechanism that through a direct reproduction procedure linked to local environment features is able to self-regulate the above exploratory swarm population, speeding it up globally.

The different experiments carry out on different sections, demonstrate that *SRS* is able for quick adaptive response, outperforming results not only with standard Genetic Algorithms, Bacterial Foraging Optimization algorithms [29] as well as comparing it with very recently successful approaches based on Co-Evolution [20]. From the different *SRS* features, we highlight some successful behaviors found: (1) *SRS* is able to maintain a number of different solutions while adapting to new peaks appearing in the landscape. (2) The experiments show that the *SRS* approach is able to spontaneously create and maintain (for certain instable changing conditions), different subpopulations on different peaks, emerging different exploratory corridors with intelligent path planning capabilities. In addition, these population split-offs only occurs when needed. For instance, in between static conditions temporal windows (when the upgrade frequency determines that the changes will only occur after some *T* time steps), the population quickly converges to a single exploratory cluster. When however, the changes arrive, and for certain environmental conditions, the swarm may split again (division of labor) in two or more subpopulations. (3) For dramatic conditions, the swarm is able to request more exploratory agents, reproducing more and scaling up its population. After this is not necessary and redundant (a highly-performing peak is found) however, the swarm can self-regulate down their number, scaling it down, thus economizing its search resources. (4) The approach allows self-organized injection of larger diversity when needed, as we can perceive from fig. 8e. (5) For certain environmental conditions, the swarm not only is able to smoothly track the dynamic extrema, as for many time-steps over some different temporal windows, is able to capture it in perfection. (6) Last but not least, we illustrate that our *SRS* collective swarm of ant-like agents is able to track about 65% of moving peaks traveling up to ten times faster than the velocity of a single individual composing that precise swarm tracking system. This emerged behavior is probably one of the most interesting ones achieved by the present work.

## Acknowledgment

The authors would like to thank *Chien-Feng Huang* (Michigan Univ.) and *Luis M. Rocha* (Indiana Univ.) for their kind attention and permission using figures 6a, 8a and 8b, respectively. We would also like to thank to Professor *Fernando Durão* (CVRM-IST, Technical Univ. of Lisbon, Portugal) for providing the MATLAB version, working in reverse communication, of the Ordinary Differential Equation (ODE) solver DOPRI853 developed and coded in FORTRAN by *E. Hairer*, *S.P.Norset* and *G.Wanner* [18]. The second author wishes to thank FCT, *Ministério da Ciência e Tecnologia - Portugal*, for his research fellowship SFRH/BD/18868/2004.

## References

[1] Angeline, P.J., "Tracking Extrema in Dynamic Environments", in Angeline, Reynolds, McDonnell and Eberhart (Eds.), Proc. of the 6[th] Int. Conf. on Evolutionary Programming, pp. 335-345, LNCS, Vol. 1213, Springer, 1997.

[2] Bäck, T., "On the Behavior of Evolutionary Algorithms in Dynamic Environments", in *IEEE Int. Conf. on Evolutionary Conference*, pp. 446-451, 1998.

[3] Bäck. T., *Evolutionary Algorithms in Theory and Practice*, Oxford University Press, 1996.

[4] Bonabeau, E., Dorigo, M., Theraulaz, G., *Swarm Intelligence: From Natural to Artificial Systems*, Santa Fe Institute series in the Sciences of Complexity, Oxford Univ. Press, NY, Oxford, 1999.

[5] Branke J., Schmeck, H., "Designing Evolutionary Algorithms for Dynamic Optimization problems", In S. Tsutsui and A. Ghosh (Eds.), *Theory and Application of Evolutionary Computation: Recent Trends*, pp. 239-262. Springer, 2002.

[6] Branke, J, *Evolutionary Optimization in Dynamic Environments*, Kluwer, 2002.

[7] Branke, J, "Memory enhanced Evolutionary Algorithms for Changing Optimization Problems", in Congress of Evolutionary Computation, CEC'99, vol. 3, pp. 1875-1882, IEEE Press, 1999.

[8] Carlisle A., Dozier G., "Tracking Changing Extrema with Adaptive Particle Swarm Optimizer", in ISSCI, World Automation Congress, Orlando, Florida, USA, June, 2002.

[9] Carlisle A., Dozier G., "Adapting Particle Swarm Optimization to Dynamic Environments", in ICAI´00, International Conference on Artificial Intelligence, Las Vegas, Nevada, USA, 2000.

[10] Chialvo, D.R., Millonas, M.M., "How Swarms build Cognitive Maps", In Steels, L. (Ed.): *The Biology and Technology of Intelligent Autonomous Agents*, 144, NATO ASI Series, 439-450, 1995.

[11] Cobb, H.G., "An investigation into the use of hypermutation as an adaptive operator in Genetic Algorithms having continuous, time-dependent nonstationary environments", Technical Report AIC-90-001, Naval Research Laboratory, Washington DC, 1990.

[12] Dasgupta D., McGregor D.R., "Nonstationary function optimization using the structured Genetic Algorithm", in Manner R., Manderick B. (Eds.), *Parallel Problem Solving from Nature*, pp. 145-154, Elsevier Science, 1992.

[13] Fernandes, C., Ramos, V., Rosa, A.C., "Varying the Population Size of Artificial Foraging Swarms on Time Varying Landscapes", in W. Duch, J. Kacprzyk, E. Oja, S. Zadrozny (Eds.), *Artificial Neural Networks: Biological Inspirations*, LNCS series, Vol. 3696, Part I, pp. 311-316, Springer-Verlag, Sept. 2005.

[14] Grefenstette J.J., "Evolvability in Dynamic Fitness Landscapes": A Genetic Algorithm approach", in *Congress on Evolutionary Computation*, CEC 99, vol. 3, pp. 2031-2038. IEEE, 1999.

[15] Grefenstette, J.J., "Genetic Algorithms for Changing Environments", in Maener R. and Manderick B. (Eds.), *Parallel Problem Solving from Nature 2*, pp. 137-144, North-Holland, 1992.

[16] Guntsch, M., Middendorf M., "Applying Population Based ACO to Dynamic Optimization Problems", in Ant Algorithms, Proc. of Third International Workshop ANTS 2002, Brussels, Belgium, Springer Verlag, LNCS 2463, pp. 111-122, 2002.

[17] Guntsch, M., Middendorf M., Schmeck H.: "An Ant Colony Optimization Approach to Dynamic TSP", in L. Spector et al. (eds.) Genetic and Evolutionary Computation Conference, San Francisco, CA: Morgan Kaufmann pages 860-867, 2001.

[18] Hairer E. Norset S.P., Wanner G., "Solving Ordinary Differential Equations: I. Nonstiff Problems", 2$^{nd}$ Edition, Springer-Verlag, Springer Series in Computational Mathematics vol. 8, 1993.

[19] Holland, O., Melhuish, C.: Stigmergy, "Self-Organization and Sorting in Collective Robotics", Artificial Life, Vol. 5, n. 2, MIT Press, 173, 1999.

[20] Huang, C.-F., Rocha, Luis M., "Tracking Extrema in Dynamic Environments using a Co-Evolutionary Agent-based Model of Genotype Edition", in Hans-Georg Beyer et al. (Eds.), GECCO'05 - *Genetic and Evolutionary Computation Conference*, pp. 545-552. ACM, 2005.

[21] Huang, C.-F., *A study of Mate Selection in Genetic Algorithms*, Doctoral dissertation. Ann Arbor, MI: University of Michigan, Electrical Eng. And Computer Science, 2002.

[22] Janson, S., Middendorf, M., "A Hierarchical Particle Swarm Optimizer for Dynamic Optimization Problems", in 1st European Workshop on Evolutionary Algorithms in Stochastic and Dynamic Environments, Coimbra, Portugal, 2004.

[23] Kennedy, J. Eberhart, Russel C. and Shi, Y., *Swarm Intelligence*, Academic Press, Morgan Kaufmann Publ., San Diego, London, 2001.

[24] Liles W., DeJong K., "The usefulness of tag bits in Changing Environments", in *Congress on Evolutionary Computation*, CEC 99, vol. 3, pp. 2054-2060. IEEE, 1999.

[25] Louis S.J., Xu Z., "Genetic Algorithms for open shop scheduling and rescheduling", in Cohen M.E., Hudson D.L. (eds.), ISCA 11$^{th}$ *Int. Conf. on Computers and their Applications*, pp. 99-102, 1996.

[26] Millonas, M.M., "Swarms, Phase Transitions and Collective Intelligence", In Langton, C.G. (Ed.): *Artificial Life III*, Santa Fe Institute, Studies in the Sciences of Complexity, Vol. XVII, Addison-Wesley, Reading, Massachusetts, 417-445, 1994.

[27] Millonas, M.M., "A Connectionist-type model of Self-Organized Foraging and Emergent Behavior in Ant Swarms", J. Theor. Biol., nº 159, 529, 1992.

[28] Parrott D., Li X., "A Particle Swarm Model for Tracking Multiple Peaks in a Dynamic Environment using Speciation", in CEC´04, Proc. of the 2004 Congress on Evolutionary Computation, p.98 - 103, IEEE Piscataway, 2005.

[29] Passino, K.M., "Biomimicry of Bacterial Foraging for Distributed Optimization and Control", *IEEE Control Systems Magazine*, pp. 52-67, June 2002.

[30] Ramos V., Fernandes C., Rosa A.C., "Societal Implicit Memory and his Speed on Tracking Extrema in Dynamic Environments using Self-Regulatory Swarms", to appear in *Journal of Systems Architecture*, Farooq M. and Menezes R. (Eds.), special issue on *Nature Inspired Applied Systems*, Elsevier, Summer 2006.

[31] Ramos V., Fernandes C., Rosa A.C., "On Ants, Bacteria and Dynamic Environments", in NCA-05, *Natural Computing and Applications Workshop*, IEEE Computer Press, Timisoara, Romania, Sep. 2005.

[32] Ramos V., Muge F., Pina P., "Self-Organized Data and Image Retrieval as a Consequence of Inter-Dynamic Synergistic Relationships in Artificial Ant Colonies", in Javier Ruiz-del-Solar, Ajith Abraham and Mario Köppen (Eds.), *Soft Computing Systems - Design, Management and Applications*, Frontiers in Artificial Intelligence and Applications series, vol. 87, pp. 500-509, IOS Press, Netherlands, 2002.

[33] Ramos, V., Almeida, F., "Artificial Ant Colonies in Digital Image Habitats: A Mass Behavior Effect Study on Pattern Recognition", In Dorigo, M., Middendorf, M., Stuzle, T. (Eds.): *From Ant Colonies to Artificial Ants* - 2$^{nd}$ Int. Wkshp on Ant Algorithms, 113-116, 2000.

[34] Rocha, Luis M., Huang, C.-F., "The Role of RNA Editing in Dynamic Environments", in ALIFE9, 9$^{th}$ *Int. Conf. on the Simulation and Synthesis of Living Systems*, Boston, Massachusetts, Sep. 12-15th 2004.

[35] Theraulaz, G., Bonabeau, E., "A Brief History of Stigmergy", Artificial Life, Vol. 5, n. 2, MIT Press, 97-116, 1999.

[36] Vavak, F., Jukes K., Fogarty, T.C., "Adaptive Combustion Balancing in Multiple Burner Boiler using a Genetic Algorithm with variable range of local search", in Bäck, T. (Eds.), 7$^{th}$ *Int. Conf. on Evolutionary Computation*, Morgan Kaufmann, pp. 719-726 1997.